\documentclass[letterpaper, 10 pt, conference]{IEEEtran}
\IEEEoverridecommandlockouts

\usepackage[letterpaper,left=.75in,right=.75in,top=.75in,bottom=.75in]{geometry}

\usepackage{epsfig}
\usepackage{amsmath}
\usepackage{multirow}
\usepackage{graphicx}
\usepackage{tabularx}
\usepackage{booktabs}
\usepackage{arydshln}
\usepackage{color}
\usepackage{subfig}
\usepackage{tikz}

\definecolor{darkgreen}{rgb}{0, 0.7, 0}
\newcommand{\greenbf}[1]{\textcolor{darkgreen}{\bf #1}}
\newcommand{\green}[1]{\textcolor{darkgreen}{#1}}

\newcommand\copyrighttext{
\centering \footnotesize \copyright 2020 IEEE. Personal use of this material is permitted. Permission from IEEE must be obtained for all other uses, in any current or future media, including reprinting/republishing this material for advertising or promotional purposes, creating new collective works, for resale or redistribution to servers or lists, or reuse of any copyrighted component of this work in other works.}
\newcommand\copyrightnotice{
\begin{tikzpicture}[remember picture,overlay]
\node[anchor=south,yshift=10pt] at (current page.south) {\fbox{\parbox{\dimexpr\textwidth-\fboxsep-\fboxrule\relax}{\copyrighttext}}};
\end{tikzpicture}
}

\title{\LARGE SemanticVoxels: Sequential Fusion for 3D Pedestrian Detection using LiDAR Point Cloud and Semantic Segmentation}
\author{Juncong Fei$^{1,2}$, Wenbo Chen$^{1}$, Philipp Heidenreich$^{1}$, Sascha Wirges$^{2}$, and Christoph Stiller$^{2}$
	\thanks{$^{1}$EE Advanced Technology, Opel Automobile GmbH, 65423 Ruesselsheim, Germany. Corresponding author email: {\tt\small juncong.fei@ext.mpsa.com}}
	\thanks{$^{2}$Institute of Measurement and Control Systems, Karlsruhe Institute of Technology (KIT), 76131 Karlsruhe, Germany}}

\begin{document}
\maketitle
\thispagestyle{empty}
\pagestyle{empty}

\copyrightnotice
\begin{abstract}
3D pedestrian detection is a challenging task in automated driving because pedestrians are relatively small, frequently occluded and easily confused with narrow vertical objects. 
LiDAR and camera are two commonly used sensor modalities for this task, which should provide complementary information. 
Unexpectedly, LiDAR-only detection methods tend to outperform multisensor fusion methods in public benchmarks. 
Recently, \mbox{PointPainting} has been presented to eliminate this performance drop by effectively fusing the output of a semantic segmentation network instead of the raw image information.
In this paper, we propose a generalization of \mbox{PointPainting} to be able to apply fusion at different levels. 
After the semantic augmentation of the point cloud, we encode raw point data in pillars to get geometric features and semantic point data in voxels to get semantic features and fuse them in an effective way.
Experimental results on the KITTI \textit{test} set show that \mbox{SemanticVoxels} achieves state-of-the-art performance in both 3D and bird’s eye view pedestrian detection benchmarks. 
In particular, our approach demonstrates its strength in detecting challenging pedestrian cases and outperforms current state-of-the-art approaches.
\end{abstract}
\section{Introduction}
\label{sec:introduction}

\begin{figure}[t] 
	\begin{center}
		\includegraphics[width=1.0\columnwidth]{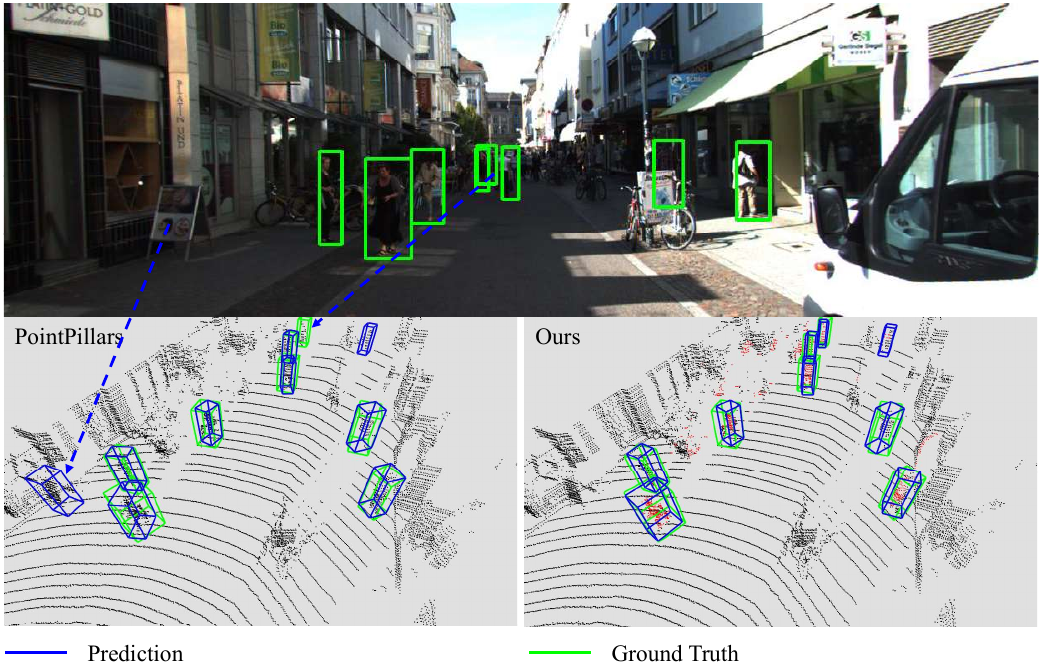}
	\end{center}
	\caption{
		Challenging pedestrian detection example from the KITTI dataset. 
		A camera image of the scene is shown at the top, the corresponding point cloud data and the pedestrian detection results of PointPillars and our method are shown at the bottom left and right, respectively. 
		Dashed arrows are used to highlight a false positive and false negative example. 
		The point cloud points corresponding to pedestrians are painted red for our method.
	Best viewed digitally with zoom.}
	\label{fig:pp_fusion}
\end{figure}

3D object detection is an important and challenging task in automated driving, in which pedestrian detection is the most difficult.
In particular, the performance and robustness of pedestrian detection is crucial for homologation and acceptance of self-driving cars. 
For 3D pedestrian detection, self-driving cars typically use LiDAR and camera sensor systems: LiDAR point clouds provide accurate spatial information, while camera images are able to provide rich contextual and semantic information. 
LiDAR-only approaches often fail to detect distant or occluded pedestrians with few points, or distinguish pedestrians from narrow vertical objects such as tall poles or outdoor signage. 
For these challenging cases, multisensor fusion becomes an obvious choice.
Despite some pioneering research in this field, such as \cite{mv3d, avod, FPointnet} or \cite{liang2018deep}, LiDAR-only methods like PointRCNN~\cite{pointrcnn} or \mbox{PointPillars}~\cite{pointpillars} present the top performance on the KITTI 3D object detection benchmark \cite{kitti}.
This is somehow counterintuitive, \textit{e.g.} as shown in Figure~\ref{fig:pp_fusion}, \mbox{PointPillars} predicts an outdoor signage as pedestrian, whereas this kind of false positive can be clearly recognized in the image, suggesting additional information from the image should be beneficial for 3D pedestrian detection using a LiDAR point cloud. 
It is reported in \cite{fusiondrop} that data from different modalities typically overfit and generalize at different rates, so that fusion may lead to a performance drop and it is of significant importance to find a proper fusion method to combine the data.

Recently, PointPainting~\cite{vora2019pointpainting} has been presented to improve the performance of LiDAR-only methods by utilizing additional semantic information from a camera image via a sequential fusion.
It projects 3D points onto the output of an image-only semantic segmentation network and uses the obtained semantic segmentation scores to paint the respective 3D points, before they are consumed in a LiDAR-only object detection network. 
However, the fusion is limited to the raw data level, which may not be optimal.

\begin{figure*}[t]
	\begin{center}
		\includegraphics[width=1.0\textwidth]{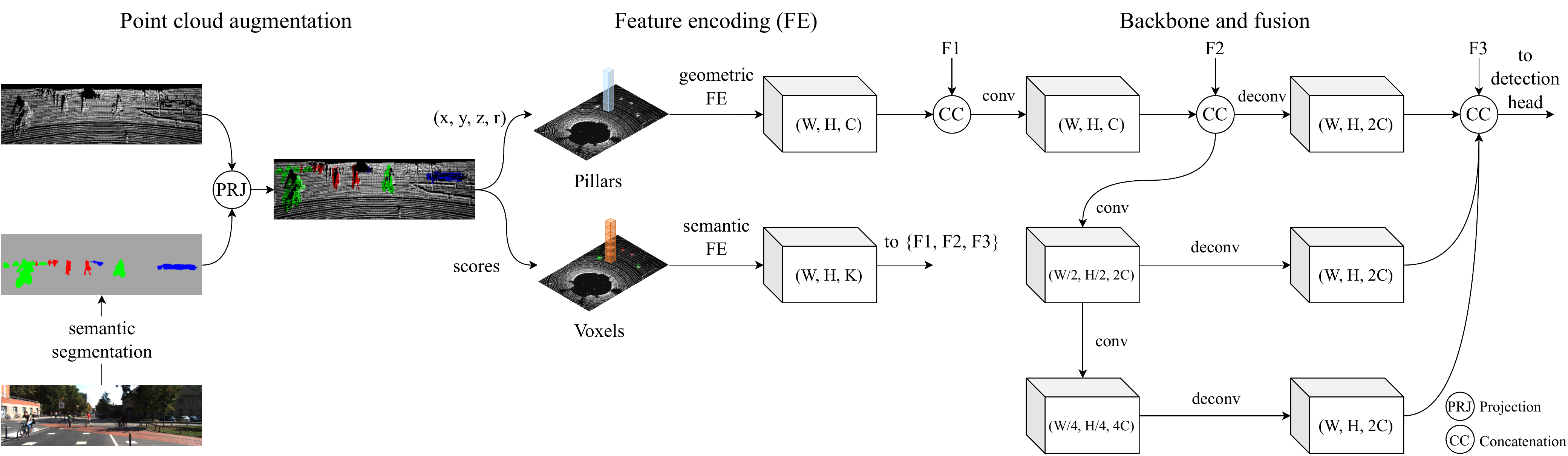}
	\end{center}
	\caption{
		Our method can be divided into four parts: Point cloud augmentation, feature encoding, backbone and detection head.
		In the first part, the point cloud is augmented with semantic information from the camera image.
		In the second part, geometric features in pillars and semantic features in voxels are both encoded in a BEV representation.
		In the backbone, geometric and semantic features are fused at three different levels $F_1$, $F_2$ and $F_3$, representing early fusion, middle fusion and late fusion, respectively. 
		The output of the backbone is then fed to the detection head, which outputs oriented 3D boxes.
	}
	\label{fig:architecture}
\end{figure*}

In this work, we propose \mbox{SemanticVoxels}: a generalization of \mbox{PointPainting} to be able to apply fusion at different levels. 
As \mbox{PointPainting}, \mbox{SemanticVoxels} appends raw point data with semantic information that has been obtained from an image-only semantic segmentation network. 
Instead of feeding the augmented point cloud to a LiDAR-only object detection network, \mbox{SemanticVoxels} further encodes semantic features in voxels and fuses them with geometric features from the original point cloud. 
In this way, we are able to apply fusion at different levels. 
In order to encode geometric features, we follow \mbox{PointPillars} to voxelize the point cloud into a set of pillars in the $xy$ plane and encode raw point data in the generated pillars.
To encode additional semantic features, the semantic point data in pillars is further voxelized along $z$ direction.

\mbox{SemanticVoxels} is a sequential fusion method since the segmentation network is independently trained. 
Despite it is not end-to-end optimized, we can still benefit from our sequential architecture.
In contrast to raw image data or middle-layer features, the output of a semantic segmentation network is a compact summary of the image, which is effective in terms of feature representation and memory efficient. 
Besides, semantic segmentation networks are widely used in autonomous vehicles and significant progress is being made in that field.
Hence, SemanticVoxels could benefit from that progress with a minimal overhead. 

We evaluate \mbox{SemanticVoxels} on the KITTI 3D and \mbox{bird’s eye view (BEV)} object detection benchmarks.
Since the pedestrian class shows worst performance in the KITTI leaderboard when compared with the car and cyclist classes, we expect most improvement potential and thus focus on the pedestrian class in this paper. However, we note that \mbox{SemanticVoxels} is also valid for cars and other traffic participants.
Experimental results on the KITTI \textit{test} set show that \mbox{SemanticVoxels} achieves state-of-the-art performance in both 3D and BEV object detection tasks. 
In particular, our approach shows its strength in detecting challenging pedestrian cases and outperforms current state-of-the-art approaches.
Further experiments on the KITTI \textit{validation} set demonstrate the effectiveness of our fusion schemes.

In summary, the main contributions of our work are:

1)  We propose a generalization of \mbox{PointPainting} to be able to apply fusion at different levels.

2)  The proposed approach encodes semantic features from the camera image in voxels and provides an effective feature representation.

3)  Our approach shows state-of-the-art performance on the KITTI 3D and BEV pedestrian detection benchmarks. Experimental results also demonstrate the superior performance of \mbox{SemanticVoxels} in detecting challenging pedestrian cases and the effectiveness of our proposed fusion schemes.

\section{Related Work}
\label{sec:related_work}

\subsection{3D Pedestrian Detection}
\label{sec:related_work_3d_pedestrian_detection}
We focus on two categories of 3D pedestrian detection methods: LiDAR-only and fusion-based methods.

\textbf{LiDAR-only.} 
PointRCNN~\cite{pointrcnn} uses a two-stage architecture to detect 3D objects, one stage for generation of 3D object proposals and one for prediction refinement.
However, the two-stage architecture achieves a better performance at the cost of more computational complexity.
\mbox{VoxelNet}~\cite{voxelnet} and \mbox{SECOND}~\cite{second} voxelize the point cloud into grid cells, then use 3D convolution for feature extraction, and 2D convolution for detection. 
Although both methods show strong performance, 3D convolutions become a bottleneck for faster inference.
\mbox{PointPillars}~\cite{pointpillars} improves the calculation speed by voxelizing the point cloud into pillars so that the 3D convolution for feature extraction is replaced by 2D convolution.
Despite the runtime improvement, it remains challenging to distinguish pedestrians from narrow vertical objects because of the lack of contextual information in point clouds.

\textbf{Fusion-based.} 
Typical two-stage detection methods \cite{mv3d, avod, feng2020leveraging} fuse extracted ROI features from the image and the projected point cloud in BEV or front view.
However, point cloud projection may cause information loss which leads to inaccurate pedestrian detections. 
ContFuse~\cite{liang2018deep} transforms extracted image features into the 2D BEV plane and fuses them with top-view features of the LiDAR stream at different scales.
Such method is limited since each BEV space corresponds to multiple pixels in the image and feature blurring might occur.
\mbox{F-PointNets}~\cite{FPointnet} and RoarNet~\cite{shin2019roarnet} first conduct 2D detection on the image to extract point clouds from the detected regions. 
Then, 3D detection is performed for each selected point cloud. 
The overall performances of both methods depend heavily on the 2D object detection since undetected objects in the image can not be recovered in later stages. 
\mbox{PointPainting}~\cite{vora2019pointpainting} and LRPD~\cite{furst2020lrpd} both use the output of an image segmentation network as auxiliary information in the detection pipeline. 
\mbox{PointPainting} complements each point with semantic information and uses LiDAR-only methods for 3D object detection. 
By design, it can only fuse features at the raw data level and thus may not be optimal. 
LRPD uses instance segmentation information to generate 3D proposals, and then conducts 3D object detection with the raw point data and the image. 
However, instance segmentation information is solely used for object proposals and not considered later.

\subsection{Semantic Segmentation}
Semantic segmentation plays an important role in scene understanding for automated driving.
As a pioneering work, FCN~\cite{fcn} adopts fully convolutional layers and combines coarse and fine features at different levels. 
The family of DeepLab~\cite{deeplabv3} uses atrous convolutions and atrous spatial pyramid pooling modules to enlarge field-of-views and improve the contextual understanding. 
BiSeNet~\cite{bisenet} improves the inference speed by designing separated paths for spatial and contextual information, but with a relatively low performance.
In this paper, we use DeepLabv3+~\cite{deeplabv3} as a trade-off between efficiency and performance.
\section{Architecture}
\label{sec:architecture}

The proposed approach, depicted in Figure~\ref{fig:architecture}, takes an image and a 3D point cloud as input and estimates oriented 3D boxes for pedestrians. 
It consists of four main blocks: point cloud augmentation, feature encoding, backbone, and detection head. We present the details of each block in the following sections.

\subsection{Point Cloud Augmentation}
\textbf{Semantic Segmentation.}
Taking an image as input, a semantic segmentation network predicts class probabilities for each pixel in the image. 
Fusion in a sequential fashion allows the use of any image segmentation network in our 3D pedestrian detection pipeline. 

\textbf{Projection and  Augmentation.}
In the KITTI dataset, each LiDAR point in the point cloud is represented by $(x, y, z, r)$, where $x$, $y$, and $z$ are the 3D coordinates and $r$ is the reflectance. 
After obtaining the semantic segmentation scores, we use them to augment the original point cloud. 
To achieve this, 3D LiDAR points are first projected onto the image plane using the homogeneous transformation matrices provided by the KITTI dataset. 
For each point, which falls in the image after transformation, the semantic segmentation scores in the corresponding pixel are then appended to the original 3D point. 
The segmentation network used in this paper outputs probabilities for four classes, which are pedestrian, cyclist, car and background. 
Thus, the points in the augmented point cloud are now $ D = 8$ dimensional.

\begin{table*}[t]
	\centering
	\caption{Performance comparison with selected state-of-the-art approaches on the KITTI test set for the \textit{Pedestrian} class. 
		Results are reported for 3D and BEV object detection tasks. 
		mAP represents the mean of APs over the three difficulty levels. 
		Our approach outperforms all listed approaches in the \textit{hard} level and in mAP metric, for both 3D and BEV.}
	\label{kitti_test}
	\begin{tabular}{|c|c||c|c|c|c||c|c|c|c|}
		\hline
		\multirow{2}{*}{Method}
		& \multirow{2}{*}{Modality} 
		& \multicolumn{3}{c|}{$\mathrm{AP}_{\mathrm{3D}}$} 
		& \multicolumn{1}{c||}{\multirow{2}{*}{$\mathrm{mAP}_{\mathrm{3D}}$}}
		& \multicolumn{3}{c|}{$\mathrm{AP}_{\mathrm{BEV}}$} 
		& \multicolumn{1}{c|}{\multirow{2}{*}{$\mathrm{mAP}_{\mathrm{BEV}}$}} \\
		\cline{3-5}\cline{7-9}
		& & Easy & Moderate & Hard & \multicolumn{1}{c||}{} & Easy & Moderate & Hard & \multicolumn{1}{c|}{} \\
		\hline
		\hline
		VoxelNet~\cite{voxelnet}  & LiDAR & 39.48 & 33.69 & 31.51 & 34.89 & 46.13 & 40.74 & 38.11 & 41.66 \\ 
		PointPillars~\cite{pointpillars}  & LiDAR & \bf{51.45} & 41.92 & 38.89 & 44.09 & 57.60 & 48.64 & 45.78 & 50.67 \\
		PointRCNN~\cite{pointrcnn}  & LiDAR & 47.98 & 39.37 & 36.01 & 41.12 & 54.77 & 46.13 & 42.84 & 47.91 \\
		AVOD~\cite{avod} & LiDAR \& Image & 36.10 & 27.86 & 25.76 & 29.91 & 42.58 & 33.57 & 30.14 & 35.43 \\
		AVOD-FPN~\cite{avod} & LiDAR \& Image & 50.46 & \bf{42.27} & 39.04 & 43.92 & 58.49 & \bf{50.32} & 46.98 & 51.93 \\
		F-PointNet\cite{FPointnet}  & LiDAR \& Image & 50.53 & 42.15 & 38.08 & 43.59 & 57.13  & 49.57 & 45.48 & 50.73 \\
		Painted PointPillars~\cite{vora2019pointpainting}  & LiDAR \& Image & 50.32 & 40.97 & 37.87 & 43.05 & 58.70  & 49.93 & 46.29 & 51.64 \\
		\hline
		Ours & LiDAR \& Image & 50.90 & 42.19 & \bf{39.52} & \bf{44.20} & \bf{58.91} & 49.93 & \bf{47.31} & \bf{52.05} \\
		\hline
	\end{tabular}
\end{table*}

\subsection{Feature Encoding}
Given the augmented point cloud, we encode geometric and semantic features in a BEV representation. 
By fusing them, the network is able to exploit geometric information from the LiDAR point cloud and semantic information from the camera image.

\textbf{Geometric Features.}
We follow \mbox{PointPillars} to encode the point cloud as pillars, where a pillar is a special voxel without extent limitation in the $z$ axis. 
First, a point cloud is voxelized into a set of pillars in the $xy$ plane. 
The original points with $(x, y, z, r)$ encoding in pillars are further decorated with the offsets from pillar point cluster center $ (\Delta x_{c}, \Delta y_{c}, \Delta z_{c})$ and the offsets from pillar center $ (\Delta x_{p}, \Delta y_{p})$. 
The dimension of resulting point encoding is thus $ D = 9$. 
The number of pillars per point cloud $(P)$ and the number of points per pillar $(N)$ are pre-defined to create a tensor with fixed size $(D, P, N)$. 
This is achieved by random sampling when there are too many points per pillar, or zero padding when there are too few.

The points in each pillar are then consumed by a simplified PointNet~\cite{pointnet}, and the output is a tensor of size $(C, P)$ with $ C = 64$. 
Each pillar is now represented by a \mbox{$C$-channel} learned feature encoding. 
Finally, all pillar features are scattered back to the pillar locations to create a BEV representation of size $(W, H, C)$, where $W$ and $H$ denote the width and height of the canvas, corresponding to $y$ and $x$ direction, respectively. 
Since the learned pillar features are mainly geometry related, we call them geometric features.

\textbf{Semantic Features.}
To obtain semantic features we further voxelize the points along $z$ direction, while maintaining the same pillar structure in the $xy$ plane. 
In this way, the semantic information at various heights is kept, which is beneficial for handling challenging cases such as occluded pedestrians or noisy segmentation results. 
The semantic features of each voxel are summarized by calculating the mean probabilities over the classes of all points in the voxel. 
This guarantees efficient utilization of image semantic information. 
Next, the voxel-wise semantic features are stacked along the $z$ axis and a $1 \times 1$ convolutional layer is applied to aggregate voxel-wise features. 
Finally, we obtain semantic features in a BEV represenation of size $(W, H, K)$ with the output channel \mbox{$K = 8$}. 

\subsection{Backbone and Fusion Schemes}
\label{fusion}
\textbf{Backbone.} 
We use the same backbone (Figure~\ref{fig:architecture}) as \mbox{PointPillars}. 
The backbone has three blocks. 
The first block keeps the size of the feature map, while the second and third block downsample the feature map by half. 
The output of every block is then upsampled and the resulting features are concatenated to create the final feature map. 

\textbf{Fusion Schemes.}
The feature fusion can be performed at different levels with various combination operations. 
Typical fusion schemes include \textit{early}, \textit{middle} and \mbox{\textit{late fusion}~\cite{mv3d,avod}}. 
\textit{Early fusion} combines information from different features in the input stage. 
In contrast, \textit{late fusion} uses separate networks to learn domain specific features from the individual inputs, followed by a combination before the prediction stage. 
\textit{Middle fusion} combines the features at intermediate layers, either once at a single layer or in a hierarchical fashion, {\it i.e.} \textit{deep fusion}. 
The combination operations can be element-wise mean or concatenation.

In this paper, we employ all three fusion schemes with concatenation operation to combine geometric and semantic features. 
\textit{Early fusion}: the features are fused before the first block and fed to the backbone network. 
\textit{Middle fusion}: semantic features are combined with the output of the first block. 
\textit{Late fusion}: semantic features are concatenated with the final feature map in the backbone.  

\subsection{Detection Head}
\textbf{3D Box Prediction.}
To predict oriented 3D boxes, we use the same detection head as in \cite{pointpillars, second}. 
It consumes the feature map from the backbone network and consists of three independent $1 \times 1$ convolution layers. 
The $1 \times 1$ convolutions are specialized for the desired tasks, namely anchor classification, box offsets regression and direction regression, respectively.

\textbf{Loss Function.}
We parameterize ground truth and anchor boxes as $ (x, y, z, l, w, h, \theta)$. 
The offsets between a ground truth box $gt$ and prior anchor box $a$ are encoded as
\begin{equation}
	\begin{aligned}
		\Delta x &= \frac{x^{gt}-x^a}{d^a}, \Delta y = \frac{y^{gt}-y^a}{d^a}, \Delta z = \frac{z^{gt}-z^a}{h^a}, \\
		\Delta l & = \mathrm{log}(\frac{l^{gt}}{l^{a}}), \Delta w = \mathrm{log}(\frac{w^{gt}}{w^{a}}), \Delta h = \mathrm{log}(\frac{h^{gt}}{h^{a}}),  \\
		\Delta {\theta}&=\theta^{gt}-\theta^a,
	\end{aligned}
\end{equation}
where $d^a = \sqrt{(l^{a})^2+(w^{a})^2}$. 
SmoothL1 is used for the 3D box regression loss $\mathcal{L}_{\mathrm{reg}}$, overall we get:
\begin{equation}
	\mathcal{L}_{\mathrm{reg}} = \sum_{b \in (\Delta x, \Delta y, \Delta z, \Delta w, \Delta l, \Delta h, \Delta \theta)} \text{SmoothL1}\left(\tilde{b} - b\right),
\end{equation}
where $\tilde{b}$ denotes the predicted box offsets. In particular, following \cite{second}, the \emph{sine} function is applied to calculate the orientation loss. 
Note that this orientation loss could not distinguish boxes when they are shifted by $\pm\pi$ radians.
To address this issue, the direction, either positive or negative, is learned with a softmax classification loss $\mathcal{L}_{\mathrm{dir}}$.

Considering the number of positive and negative samples is imbalanced, focal loss~\cite{lin2017focal} is used as the object classification loss:
\begin{equation}
	\mathcal{L}_{\mathrm{cls}}= -\alpha_\mathrm{t} \left(1 - p_\mathrm{t}\right) ^ \gamma \log p_\mathrm{t} ,
\end{equation}
where $p_\mathrm{t}$ indicates the class probability of an anchor.
During training, we use the recommended settings in \cite{pointpillars, lin2017focal} with $\alpha=0.25$ and $\gamma=2$.
The total loss is therefore defined as:
\begin{equation}
	\mathcal{L}_{\mathrm{total}} = \frac{1}{N_{\mathrm{pos}}}\left(\beta_{1} \mathcal{L}_{\mathrm{reg}} + \beta_{2} \mathcal{L}_{\mathrm{dir}}  + \beta_{3} \mathcal{L}_{\mathrm{cls}}\right),
\end{equation}
where $N_{\mathrm{pos}}$ denotes the number of positive anchors. 
$\beta_{i}$ are constant weights for balancing different losses. 
We set \mbox{$\beta_{1}=2.0$}, $\beta_{2}=0.2$, and $\beta_{3}=1.0$.

\begin{table*}[t]
	\centering
	\caption{Performance comparison among the different fusion schemes and PointPillars on the KITTI \textit{validation} set. 
		Results are reported for 3D and BEV object detection tasks. 
		Compared with the reproduced results of PointPillars, our fusion approaches show a significant improvement on all comparisons, indicating the importance of semantic features in 3D pedestrian detection.
		\textit{Early fusion} achieves the best mAP improvement of 3.2 for 3D and 2.82 for BEV detection tasks. 
	}
	\label{kitti_val}
	\begin{tabular}{|c||c|c|c|c||c|c|c|c|}
		\hline
		\multirow{2}{*}{Method}
		& \multicolumn{3}{c|}{$\mathrm{AP}_{\mathrm{3D}}$} 
		& \multicolumn{1}{c||}{\multirow{2}{*}{$\mathrm{mAP}_{\mathrm{3D}}$}}
		& \multicolumn{3}{c|}{$\mathrm{AP}_{\mathrm{BEV}}$} 
		& \multicolumn{1}{c|}{\multirow{2}{*}{$\mathrm{mAP}_{\mathrm{BEV}}$}} \\
		\cline{2-4}\cline{6-8}
		& Easy & Moderate & Hard & \multicolumn{1}{c||}{} & Easy & Moderate & Hard & \multicolumn{1}{c|}{} \\
		\hline
		\hline
		PointPillars   & 67.69 & 60.94 & 56.02 & 61.55 &73.42 & 68.89 & 63.32 & 68.54 \\ 
		\hline
		Late Fusion   & 68.58 & 62.62 & 59.19 & 63.46 & 73.72 & 69.86 & 64.29 &69.29 \\ 
		Delta  		 & \green{+0.89} & \green{+1.68} & \green{+3.17} & \green{+1.91} & \green{+0.30} & \green{+0.97} & \green{+0.97} & \green{+0.75} \\ \hdashline
		Middle Fusion & 68.15 & 63.31 & 59.15 & 63.54 & 74.31 & 70.81 & \bf{66.54} & 70.55 \\
		Delta  		 & \green{+0.46} & \green{+2.37} & \green{+3.13} & \green{+1.99} & \green{+0.89} & \green{+1.92} & \greenbf{+3.22} & \green{+2.01} \\ \hdashline
		Early Fusion  & \bf{69.71} & \bf{64.47} & \bf{60.07} & \bf{64.75} & \bf{76.31} & \bf{71.56} & {66.21} & \bf{71.36} \\
		Delta  		 & \greenbf{+2.02} & \greenbf{+3.53} & \greenbf{+4.05} & \greenbf{+3.20} & \greenbf{+2.89} & \greenbf{+2.67} & \green{+2.89} & \greenbf{+2.82} \\
		\hline
	\end{tabular}
\end{table*}
\section{Experimental Setup}
\label{sec:experimental_setup}

We evaluate our approach on the KITTI 3D object detection benchmark~\cite{kitti} focusing on the {\it Pedestrian} class.

\subsection{Dataset and Evaluation Metric}
The KITTI dataset contains 7,481 images and point clouds for training and 7,518 for testing, covering three categories: {\it Car}, {\it Pedestrian}, {\it Cyclist}. 
For experimental evaluation, we follow \cite{mv3d, avod} to split the training samples into 3,712 training and 3,769 validation samples. 
When evaluating on the test set, we follow \cite{pointpillars} to perform training on 6,733 samples from the training samples, and evaluation on the rest.

The results are evaluated using Average Precision (AP) for both 3D and BEV evaluations with an Intersection over Union (IoU) threshold of 0.5 for the {\it Pedestrian} class. 
The evaluation is conducted in three difficulty levels, namely {\it easy}, {\it moderate} and {\it hard}, according to the occlusion level, maximal truncation and the height of the 2D box in image.

\subsection{Implementation Details}
We use DeepLabv3+~\cite{deeplabv3} for semantic segmentation. 
The network is pretrained on the Cityscapes dataset~\cite{cordts2016cityscapes} and finetuned on the KITTI semantic segmentation dataset. 
When using the KITTI detection dataset, we keep the probabilities for pedestrians, cyclists and cars. 
The probabilities of background are calculated by subtracting the probabilities of the above three classes from one.

For all experiments of the 3D object detection network, we follow \cite{pointpillars} to crop the point cloud at $[(0, 48), (-20, 20), (-2.5, 0.5)]$ meters along $x$, $y$, $z$ axes respectively.
When generating pillars, we use a pillar grid size of $0.16^2\,\mathrm{m}^2$, number of pillars $P=12000$, and number of points per pillar $N=100$. 
The voxels for encoding semantic features have a $z$ resolution of 0.3 meters. 
For the pedestrian anchor, we use the same size as \mbox{PointPillars}, which is $(0.6, 0.8, 1.73)$ meters in width, length and height, respectively. 
The anchor is applied at two angles: 0 and 90 degrees and the center is located at -0.6 meters in $z$ axis.

The detection network is trained using the Adam optimizer with an initial learning rate 0.0002. 
We apply exponential decay to the learning rate with a factor of 0.8 every 15 epochs.
The training lasts 160 epochs with a mini-batch size of 2 and 4 on the \textit{validation} and \textit{test} set, respectively.
\section{Results}
\label{sec:results}

\subsection{Quantitative Analysis}
\textbf{Evaluation on the KITTI Test Set.} 
The experimental results on the KITTI \textit{test} set are presented in Table~\ref{kitti_test}. 
We compare our approach with state-of-the-art methods in both 3D and BEV object detection tasks.
All metrics are calculated using an IoU threshold of 0.5.
The mean average precision (mAP) over the three difficulty levels is computed to represent the overall performance.
Our approach outperforms the competing approaches in terms of 3D mAP and BEV mAP, achieving state-of-the-art performance.
In addition, we achieve competitive results across all difficulty levels in both 3D and BEV. 
In particular, a higher performance in the \textit{hard} level demonstrates that the semantic information is of great importance for detecting hard cases,  such as occluded or distant pedestrians. 
Note that these cases are of big concern in automated driving, and we will discuss them in details in Section~\ref{qualitative}. 

\textbf{Evaluation on the KITTI Validation Set.}
To analyze the importance of semantic features and the effectiveness of different fusion schemes, further experiments on the KITTI validation set are conducted. 
We use \mbox{PointPillars} as baseline and implement our approach using three different fusion schemes, namely \textit{early}, \textit{middle} and \textit{late fusion}, as introduced in Section~\ref{fusion}. 
All networks are evaluated in 3D and BEV object detection tasks by AP and mAP on the KITTI \mbox{\textit{validation}} set. In addition, we compute the difference between the fusion approach and the baseline. 
Table~\ref{kitti_val} summaries the evaluation results.
When comparing with the baseline, our fusion approaches show a significant improvement on all comparisons, indicating the importance of semantic features in 3D pedestrian detection.
We further compare APs in \textit{easy}, \textit{moderate} and \textit{hard} levels and found that semantic features lead to the largest boost in the \textit{hard} level. 
This is consistent with what we observed on the KITTI \textit{test} set, and provides further evidence that shows the benefit of leveraging semantic information from image for 3D pedestrian detection. 
Among three fusion schemes, \textit{early fusion} achieves the best mAP improvement of 3.20 for 3D and 2.82 for BEV. 
This suggests that the used backbone network is powerful enough to combine the geometric and semantic features in a proper way. 

\subsection{Qualitative Analysis}
\label{qualitative}
\begin{figure*}[t]
	\begin{center}
		\includegraphics[width = 1.0\textwidth]{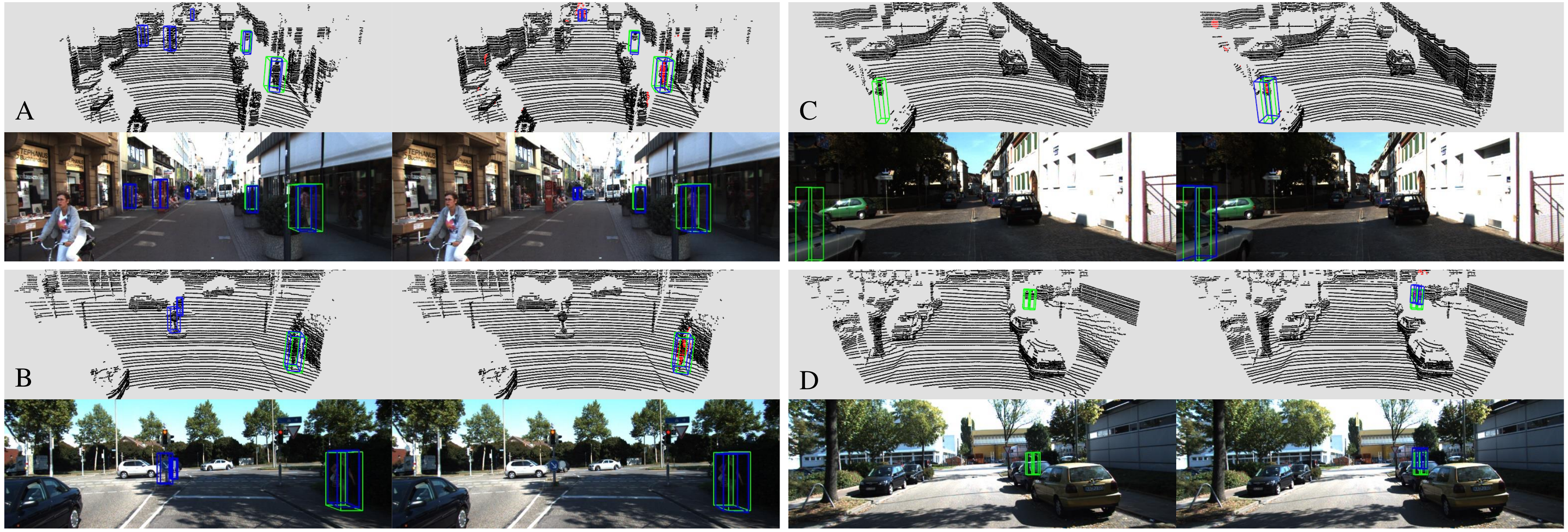}
	\end{center}
	\caption{Qualitative results produced by PointPillars and our early fusion approach on the KITTI dataset. 
		Ground truth boxes are displayed in green and predictions in blue.
		For each comparison, the left column shows the results of PointPillars in the point cloud (upper) and image (lower), while the right column shows the results of our approach.
		A-B: PointPillars predicts a postcard rack, a shelf and two road signs as pedestrian, while our approach handles these cases correctly. 
		\mbox{C-D: Our approach} successfully detects severely occluded pedestrians, while PointPillars fails.
		Best viewed digitally with zoom.}
	\label{fig:kitti_samples}
\end{figure*}

We show a qualitative comparison between \mbox{PointPillars} and our \textit{early fusion} approach in Figure~\ref{fig:kitti_samples}. 
Figures~\ref{fig:kitti_samples}-A and \ref{fig:kitti_samples}-B show common failures of \mbox{PointPillars} as a LiDAR-only network struggles to distinguish between pedestrians and narrow vertical objects such as tall poles or outdoor signage. 
By incorporating additional semantic information from the image, these false positives can be easily eliminated. 
We observe from Figures~\ref{fig:kitti_samples}-C and \ref{fig:kitti_samples}-D that \mbox{PointPillars} performs poorly in detecting severely occluded pedestrians. 
Cases shown in these figures are challenging, but often appear in daily urban scenarios, \textit{e.g.} pedestrians may walk out suddenly in front of parked cars. 
It is of great importance to detect them at an early stage, even when parts of the body are not yet visible. 
The consistent detection results of our fusion approach demonstrates its effectiveness and the importance of semantic information for pedestrian detection.

\section{Conclusion}
\label{sec:conclusion}

In this paper, we presented a novel fusion method exploiting complementary strengths of LiDAR point clouds and camera images for accurate 3D pedestrian detection. 
In particular, we exploit geometric information from the LiDAR point cloud and semantic information from the camera image and fuse them in an effective way.
After point cloud augmentation using image segmentation scores, we encode these summarized image features in voxels.
In this way, semantic features are able to be fused with geometric features learned from the point cloud at different levels. 

Experiments on the KITTI object detection benchmark show that our method outperforms existing state-of-the-art methods in both 3D and BEV object detection tasks. 
Moreover, extensive experimental results on the validation set further demonstrate the effectiveness of our method in detecting challenging pedestrian cases.

In the future, we plan to investigate the dependency of our method on the quality of the semantic segmentation output. 
Moreover, we aim to study the robustness of our fusion method against adversarial attacks. Finally, we want to test our method on an experimental vehicle.

\section*{Acknowledgment}
\label{sec:acknowledgment}

The research leading to these results is funded by the German Federal Ministry for Economic Affairs and Energy
within the project ``KI-Absicherung -- Safe AI for automated driving''.
We thank Ahmed Hammam, Lukas Stäcker, and Patrick Feifel for the suggestions and discussions.

\bibliographystyle{IEEEtran}
\bibliography{root}

\end{document}